\documentclass[10pt,twocolumn,letterpaper]{article}

\usepackage{iccv}
\usepackage{times}
\usepackage{epsfig}
\usepackage{graphicx}
\usepackage{amsmath}
\usepackage{amssymb}
\usepackage{booktabs}
\usepackage{float}
\usepackage[ruled,linesnumbered]{algorithm2e}

\usepackage[breaklinks=true,bookmarks=false]{hyperref}

\iccvfinalcopy 

\def\iccvPaperID{****} 

\ificcvfinal\fi
\begin{document}

\title{Meta Self-Learning for Multi-Source Domain Adaptation: A Benchmark}

\author{Shuhao Qiu, Chuang Zhu\thanks{the corresponding author: Chuang Zhu (czhu@bupt.edu.cn)}, Wenli Zhou\\
Beijing University of Posts and Telecommunications, Beijing 100876, China\\
{\tt\small \{qiushuhao, czhu, zwl\}@bupt.edu.cn}\\
\textcolor{red}{\url{https://bupt-ai-cz.github.io/Meta-SelfLearning}}
}

\maketitle
\ificcvfinal\thispagestyle{empty}\fi

\begin{abstract}
   In recent years, deep learning-based methods have shown promising results in computer vision area. However, a common deep learning model requires a large amount of labeled data, which is labor-intensive to collect and label. What's more, the model can be ruined due to the domain shift between training data and testing data. Text recognition is a broadly studied field in computer vision and suffers from the same problems noted above due to the diversity of fonts and complicated backgrounds. In this paper, we focus on the text recognition problem and mainly make three contributions toward these problems. First, we collect a multi-source domain adaptation dataset for text recognition, including five different domains with over five million images, which is the first multi-domain text recognition dataset to our best knowledge. Secondly, we propose a new method called Meta Self-Learning, which combines the self-learning method with the meta-learning paradigm and achieves a better recognition result under the scene of multi-domain adaptation. Thirdly, extensive experiments are conducted on the dataset to provide a benchmark and also show the effectiveness of our method. The code of our work and dataset are available soon at \url{https://bupt-ai-cz.github.io/Meta-SelfLearning/}.
\end{abstract}

\section{Introduction}

In recent years, the booming development of deep learning leads to great progress in computer vision field. Text recognition has always been an important field in computer vision, for texts are everywhere in daily life and the understanding of them can be very meaningful. However, to realize accurate recognition in the real scene (which is known as scene text recognition) is still a challenging field because of the diversity of fonts and complicated environment (distortion, variation of font, occlude, etc.). Many deep-learning-based methods are proposed over the past few years to solve this problem \cite{aster, aster0, shi2016robust, wang2020decoupled,gtc}. 

As a data-driven method, the performance of the deep learning model highly relies on the amount of training data. The common way for addressing the above problem is to build publicly available large-scale datasets. However, collecting and labeling a large amount of real scene text data can be a time-consuming and labor-intensive work. What's more, the direct use of these datasets can not produce good results sometimes because of the distribution shift between the training data (source domain) and the testing data (target domain). Domain adaptation is a research field that focuses on aligning the source domain and target domain and thus obtaining better results. In recent years, many domain adaptation methods \cite{deepdomainconfusion, deepcoral, long2015learning} have been proposed to solve the domain shift problem in image classification problems. Some researches in the text recognition area are also proposed that have good results in different domains \cite{s2sda, wirteradp, gadan}. 

While single-source domain adaptation is widely researched, multi-source domain adaptation is actually more suitable for the scene text recognition problem, for the training data of scene texts are always collected from many different sources. As a generalization form of single-domain, multi-source domain adaptation universally gets a better result than single-source domain adaption for the larger training corpus. Most works in this area focus on image classification problems \cite{advermsd, momentmatchmsd}, and some datasets are proposed for this field \cite{momentmatchmsd, pacs}. However, to our best knowledge, there are no publicly available datasets for text recognition in this area, and therefore, there is almost no related research work. In this paper, we collect a multi-source domain adaption dataset and provide a benchmark to fill this gap. 

Some recent studies combined multi-source domain adaptation methods and meta-learning together. Hieu \etal \cite{metapsuedolabel} proposed a self-learning method combined with meta-learning, which achieved SOTA on many image classification tasks. Li \etal \cite{onlinemeta} proposed a framework to fit any domain adaptation methods and got better results for the good initialization provided by the meta-learning method. However, during the meta-update, this method didn't make use of the information from the target, which is very important for unsupervised domain adaptation problems. Inspired by this work, we proposed a method called meta self-learning in this paper. Our method adequately utilizes the information of the target domain by adding the target domain data to the meta-update process and get pseudo-labels with higher quality. The main contributions of our work are summarized as follows:

\begin{itemize}
\item We collect a multi-source domain adaptation dataset for text recognition with over 5 million images from 5 different domains. To the best of our knowledge, this is the first multi-domain adaptation dataset for text recognition.

\item We propose a new self-learning framework for multi-source domain adaptation, which is effective and can be easily fit into any MDA and self-learning problem. 

\item Experiments are conducted on our dataset, which provide a benchmark and show the effectiveness of our method.
\end{itemize}


\section{Related Works} \label{section:2}

\subsection{Text Recognition}
A common text recognition system can be divided into four stages: image preprocessing stage, feature extraction stage, sequence modeling stage, and prediction stage. The preprocessing stage mainly focuses on normalizing the image and rotating the text images into an appropriate position; STN \cite{stn} is a commonly used method in the computer vision area. Recent works \cite{shi2016robust, liu2016star, aster} proposed methods like thin-plate spline (TPS) transformation to get a better result. In the feature extraction stage, CNN is used to extract the generate feature maps for images. In the field of text recognition, images are always be transformed into a feature map with a height of 1, therefore can be processed as a sequence in the following stages. In the sequence modeling stage, models like Bi-LSTM or GRU are used to learn the sequence information from the feature extracted in the last stage. Due to the variable length of the data, connectionist temporal classification (CTC) \cite{ctc} or attention mechanism \cite{attnbahdanau,attnchorowski} are commonly used in the prediction stage to make the final prediction.

\subsection{Domain adaptation} 
The original domain adaptation only focuses on the single-source domain problem, and the main idea is to align the distribution between the source domain and target domain. Multi-source domain adaptation problem is also getting popular in recent years for it is more closer to the real scene and can get a better result than single-source domain adaptation generally. The main domain adaptation methods can be mainly classified into three types.

\textbf{Discrepancy based domain adaptation:}
Tzeng \etal \cite{deepdomainconfusion} proposed a domain confusion loss by calculating the maximum mean discrepancy (MMD) between the source domain data and the target domain data.
Long \etal \cite{long2015learning} proposed to calculate the MMD of more than one layer and used a multi-kernel MMD (MK-MMD) to achieve a better alignment. 
Sun \etal \cite{deepcoral} proposed CORAL loss to align the second-order statistics of the source and target distributions.
Peng \etal \cite{momentmatchmsd} proposed a multi-source domain adaptation method to calculate the moment distance not only between the source domain and target domain, but also among source domains. 

\textbf{Adversarial training based domain adaptation:}
Ganin \etal \cite{backprop} used a domain discriminator and proposed a gradient reversal layer to separate the feature extractor and the domain discriminator, which forces the feature extractor to extract the domain-invariant feature.
Zhao \etal \cite{advermsd} proposed an adversarial method to solve the multi-source domain adaptation problem using the gradient reversal layer and provided a thorough analysis.

\textbf{Self-training-based domain adaptation:} Self-training method has been widely used in image classification and segmentation problems. The method trains the model iteratively by generating pseudo-label of target data and adding them into the training data \cite{triguero2015self}. However, the direct use of this mechanism may only be helpful for the easy class and lead to a bias among classes in classification problems. Zou \etal \cite{confreg} proposed a confidence regularized self-training method by adding regularizers to the network and achieve a better result.


\subsection{Meta-Learning}
Meta-learning, also known as learning-to-learn, is a broadly studied field in recent years. Different from traditional learning methods focusing on a specific task, meta-learning methods aim to learn ``how to learn” from multiple tasks and achieve fast adaptation on a new task with a few samples.

MAML \cite{MAML} is a very famous meta-learning algorithm, which aims to learn a good initialization of parameters and can guarantee a fast convergence to local minimal with a small amount of data on a new task. However, the computation overhead of MAML during training is very high due to the calculation of second-order derivatives. To reduce the computational cost of MAML, Reptile \cite{reptile} provides a family of first-order meta-learning algorithms to approximate the original MAML. Some meta-learning methods are designed based on metric learning, such as matching network and prototypical network \cite{prototypical, matching_network}, which also make great influence in the field of few-shot learning.

Due to the inherent ``bi-level update” property \cite{meta_survey}, some meta-learning-based domain adaptation and domain generalization methods were proposed in past few years. Li \etal \cite{metadg} proposed a domain generalization method by dividing the source domains into meta-train {domains} and meta-test domains to simulate the real training process, which achieves better results on the real target domain. An online meta-learning method was proposed to enhance the effectiveness of any domain adaptation method \cite{onlinemeta}. The online meta-learning paradigm also enables the long-term effect of meta-learning, instead of only being effective at the beginning of the training stage. 

\subsection{Self-Learning}
Self-learning methods predict labels for the unlabeled data using the model trained on source domains and take them as correct labels if the predict confidence is higher than a threshold \cite{pseudolabel}. The self-learning method can always bring considerable improvement because of the direct use of target domain data. However, there also exist some problems. The generated pseudo-labels can be noisy sometimes, and lead the model to a bad local minimal. Therefore, most works focus on how to generate pseudo-labels with high quality. Recent works \cite{cbst, confreg} provide methods for balancing different classes or adding regularizers to the model. Hieu \etal \cite{metapsuedolabel} provided a meta-learning paradigm combining with the teacher-student model, thus the parameters of the teacher model can be evaluated by the pseudo-label and get updated with better quality. In this paper, we provide a new way to combine the meta-learning paradigm and the self-learning methods on the basis of \cite{onlinemeta}, by utilizing the information of pseudo-label during the meta-update.

\section{Multi-Domain Text Recognition Dataset}
In this section, we will introduce the details of our dataset. Our multi-domain text dataset consists of 5,209,215 images in total, and is divided into five domains, which are synthetic domain, handwritten domain, document domain, street view domain, and car plate domain. The character set size is set to 3,816, with 3,754 common Chinese characters and 62 alphanumeric characters, which can be represented as $C$. All of the five different domains are split into the training set and the test set with the ratio of $9:1$. The details of different domains are shown as follows.

\textbf{Synthetic Domain.} A good deep-learning-based text recognition model requires millions of images, which is very hard to collect and label. To fill up the gap of data size, synthetic texts are widely used during the training procedure. Therefore, a synthetic domain is necessary for the multi-domain dataset.

Our synthetic dataset contains 1,110,620 images in total. We generate all the training data with 5 different fonts and three different backgrounds. Random blocks and lines are added for the augmentation. When generating corpora, we found it impossible to cover all the characters in the charset only using real corpus. To achieve a better result on uncommon characters, we sacrifice the semantic information of this domain and generated all samples by random sampling from the $C$. The length of each corpus is between 4 and 10. 

\textbf{Document Domain.}
The data of document domain is collected from an open-source project\footnote{https://github.com/YCG09/chinese\_ocr}, and the dataset contains about 3 million images. We filtered out the images that contains characters not in $C$ and got 1,710,885 images in total. The corpora in this domain are from documents and news, and have the same length of 10.

\textbf{Street View Domain.}
There are many publicly available street view datasets on the internet, however, most of them only contain street view text images from one region, which means only contains Chinese character or alphanumeric characters. In order to make a better recognition result on both Chinese and alphabetical characters, we merged the images from both Chinese scene text recognition datasets and English scene text recognition datasets, including SVT \cite{svt}, SVT perspective \cite{svtp}, ICDAR2013 \cite{icdar13}, ICDAR2015 \cite{icdar15}, RCTW17 \cite{icdar17}, ICDAR-2019 \cite{icdar19}, and CUTE80 \cite{cute80}. After the same filtering operation with the document domain, we got 199,346 images in this domain.

\textbf{Handwritten Domain.}
The data of the handwritten domain is generated using the images in CASIA Online and Offline Chinese Handwriting Databases \cite{liu2011casia}. Out of the same consideration in the synthetic domain, we think that a better coverage of the charset is more important. Therefore, we use the same corpora with the synthetic domain. What's more, we also use some corpora from the street view domain to balance between the semantic information and the coverage of the charset. Images are generated by concatenating single-char images together according to the corpora. We get 1,897,021 images in total for handwritten images.

\begin{figure*}[h]
\begin{center}
\includegraphics[scale=0.5]{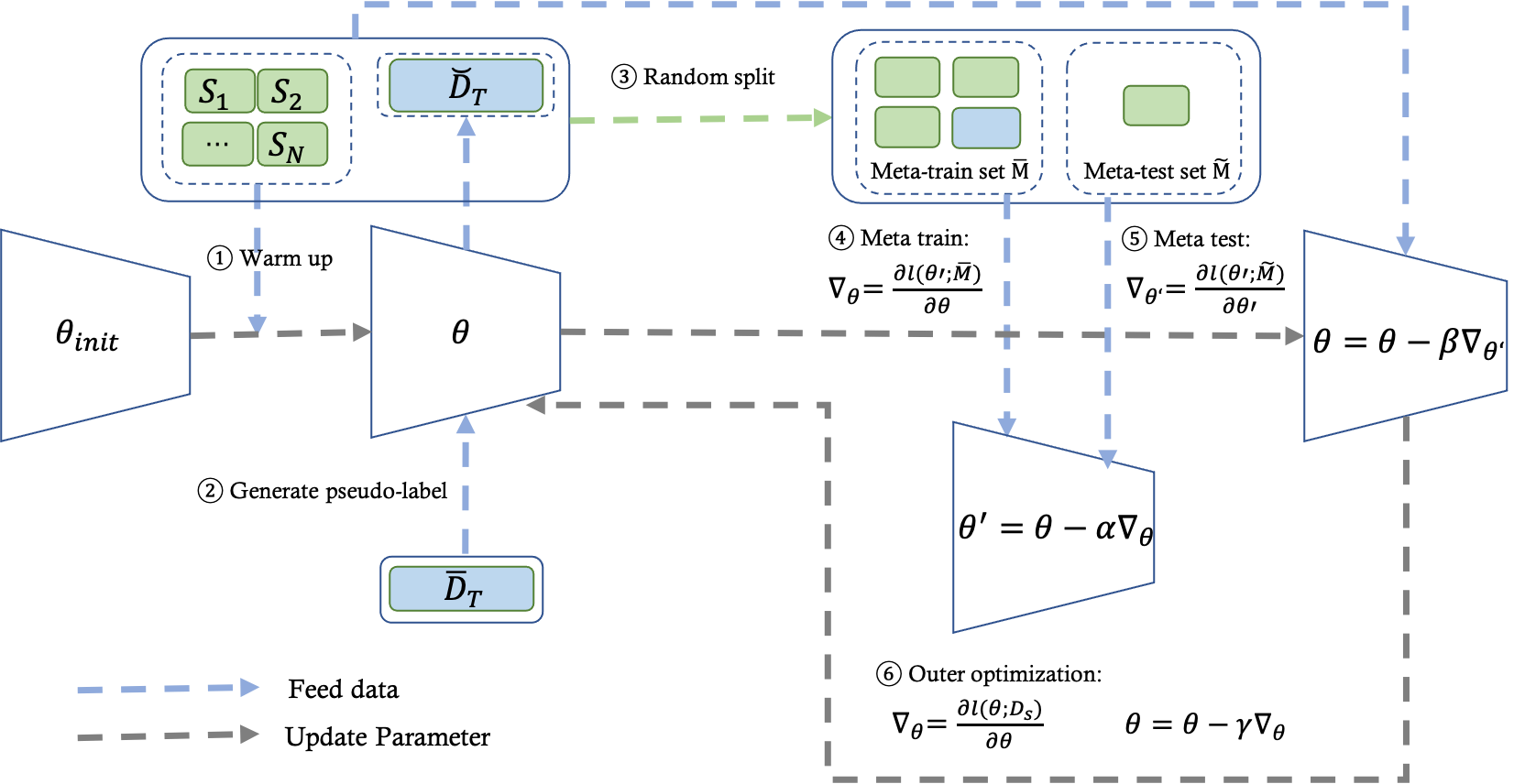}
\end{center}
\caption{Illustration of Meta Self-Learning method: The procedure of our method can be summarized as follow steps: 1. The data from source domains with labels $D_S$ are used for warm-up; 2. The model is evaluated on the target domain data without labels ${\overline{D}}_T$ and generates pseudo-labels; 3. The target domain data with pseudo-labels $D_S$ and $\breve{D}_T$ are split randomly as ${\overline{M}}$ and $\widetilde{M}$; 4. Meta train using ${\overline{M}}$; 5. Meta test using $\widetilde{M}$; 6. Outer optimization using a subset of $D_S$ and $\breve{D}_T$.}
\label{fig:methodoverview}
\end{figure*}

\textbf{Car License Domain.}
The car license domain is composed of two parts. The first part is the largest Chinese car license dataset CCPD \cite{ccpd}, and we only use the base part of this dataset, which contains 199,996 images. Although the CCPD contains a large amount of data, there exists a severe problem in this dataset. A Chinese license plate consists of 7 characters, the first one is the Chinese abbreviation of the province, and the remaining six are letters or numbers. Among the 7 characters, the abbreviation of the province is the most difficult to recognize for the Chinese characters are more complicated than alphabetic characters. However, most images in this dataset are collected from the same city, which means most of the images have the same province identity. This situation leads to a severe imbalance of the dataset, and the model being trained on this dataset can not get good performance in recognizing the province identity of other provinces. To solve this problem, we provide extra 7,932 images collected from surveillance cameras in 26 different provinces and alleviate this problem. We finally got 207,928 images in total, including 31 Chinese characters and 34 alphabetic characters. 

\section{Method}
In this paper, we focus on the problem of multi-source domain adaptation and provide a new method combining the self-learning method with the online meta-learning method. The overview of our method is shown in Fig. \ref{fig:methodoverview}. Given $D_S = \{S_1, S_2,\dots,S_N\}$ as the data with labels from multiple source domains, ${\overline{D}}_T$ as the target domain without labels, our goal is to get a model $f(\cdot)$ with parameters $\theta$ that achieves good results on the target domain using $D_S$ and ${\overline{D}}_T$. By using the self-learning method, pseudo-label will be generated using ${\overline{D}}_T$, the data with pseudo-label can be represented as ${\breve{D}}_T$. While previous work using online meta-learning method didn't take pseudo-label into account during the meat-update,  we add ${\breve{D}}_T$ into the meta-train set and the model will be updated with both $D_S$ and ${\breve{D}}_T$ during the meta-update. This setting brings great gain for the model, for the information of the target domain can be very valuable. What's more, pseudo-labels with higher quality can be acquired under this paradigm. In the following section, we will first introduce the detail of our proposed method, and then introduce the text recognition model we used.

\begin{algorithm}[h]
\caption{Meta Self-Learning for Multi-source Domain Adaptation}
\label{alg:MDS}
\KwData{$D_S=\{S_1, S_2,\dots,S_N\}, \overline{D}_T=T_1$}
\KwIn{Initial model $f(\theta)$}
\KwIn{Meta train learning rate $\alpha$, meta test learning rate $\beta$, outer learning rate $\gamma$}
\KwIn{pseudo-label threshold $\tau$}
\KwResult{Optimized parameter $\theta$}
Warming up using $D_s$, get $\theta$\;
 \While{not converge}{
  $\breve{D}_T= f(\theta;T_1) > \tau$\;
  \textbf{Random Split}: $D_S+ \breve{D}_T$ $\rightarrow$  $\overline{M} + \widetilde{M}$ \;
  Meta train: evaluate ${\nabla}_{\theta} = \frac{\partial l(\theta;\overline{M})}{\partial \theta}$\;
  Update: $\theta ' = \theta - \alpha{\nabla}_{\theta}$\;
  Meta test: evaluate $\nabla_{\theta'} = \frac{\partial l(\theta';\widetilde{M})}{\partial \theta'}$\;
  Update $\theta = \theta - \beta\nabla_{\theta^{'}}$\;
  Outer optimization:  evaluate ${\nabla}_{\theta} = \frac{\partial l(\theta;D_S)}{\partial \theta}$\;
  Update $\theta = \theta - \gamma\nabla_\theta$\;
 }
\end{algorithm}

\subsection{Meta Self-Learning}
The whole procedure of our meta self-learning method is described in Algorithm \ref{alg:MDS}. 

\textbf{Warm-Up and Generate Pseudo-Label.} The model will first be trained on $D_S$ as the warm-up phase. Warm-up is a necessary process for the self-learning method, and this process will greatly improve the quality of the generated pseudo-label and lead to a better result. Without warm-up process, the generated pseudo-labels will either have low confidence or wrong content, which will greatly jeopardize the predict accuracy on the target domain. After the warm-up, the target data with pseudo-label $\breve{D}_T$ will be generated. 

\textbf{Random Split.} The usage of the pseudo-label is one of the most important issue. As the raw pseudo-label can be noisy, a meta-update is used in our method. During the meta-update, both $D_S$ and $\breve{D}_T$ will be used, and are divided randomly into meta-train set $\overline{M}$ and meta-test set $\widetilde{M}$, which corresponds to the support set and query set in vanilla MAML. 

\textbf{Meta-Train.} The network will first update on the meta-train set $\overline{M}$ using the text recognition loss $l$ in Eq. \ref{textloss}
\begin{equation}
    l_a = \frac{1}{||\overline{M}||}\sum_{i=0}^{||\overline{M}||}l(\theta;\hat{y^i}, y^i),  \label{meta_train}
\end{equation}
where $||\overline{M}||$ is the size of meta-train set, $\hat{y^i}$ and $y^i$ are the predicted label and the ground-truth label of text image, respectively. Then, the gradient of the model parameter $\theta$ is calculated as $\nabla_{\theta}$, where 
\begin{equation}
\begin{aligned}
\nabla_{\theta}=\frac{\partial l_{a}(\theta)}{\partial \theta}.
\end{aligned}
\end{equation}
Then, the parameter will be updated as 
\begin{equation}
\begin{aligned}
\theta'=\theta - \alpha\nabla_{\theta}, 
\end{aligned}
\end{equation}
where $\alpha$ is the learning rate for the meta-train phase.

\textbf{Meta-test.} The meta-test phase is used to evaluate the model using meta-test set $\widetilde{M}$. In this phase, the loss function $l_b$ is calculated with parameter updated in meta-train phase $\theta'$.
\begin{equation}
    l_b = \frac{1}{||\widetilde{M}||}\sum_{i=0}^{||\widetilde{M}||}l(\theta';\hat{y^i}, y^i), \label{meta_test}
\end{equation}
where $||\widetilde{M}||$ is the size of meta-test set.
Following the vanilla MAML, we need to calculate the gradient of the original parameter $\theta$ using $l_b$
, which is 
\begin{equation}
\begin{aligned}
\frac{\partial l_{b}(\theta')}{\partial \theta}&=\frac{\partial l_{b}(\theta')}{\partial \theta'}\cdot\frac{\partial \theta'}{\partial\theta}
\\&=\frac{\partial l_{b}(\theta')}{\partial \theta'}\cdot\frac{\theta - \alpha\frac{\partial l_{s}(\theta)}{\partial \theta}}{\partial\theta}
\\&=\frac{\partial l_{b}(\theta')}{\partial \theta'}\cdot(1-\alpha\frac{\partial^2 l_{s}(\theta)}{\partial \theta^2})
\label{second-order}
\end{aligned}
\end{equation}

It can be seen that a second-order derivative needs to be calculated. However, calculating the second-order derivative can be prohibitively expensive for a deep learning framework, especially for a large model with a long computation graph. Therefore, a first-order approximation of MAML is widely used, we can simply neglect the second-order entry in Eq. \ref{second-order}, and the gradient $\nabla_{\theta}$ can be approximated as
\begin{equation}
\begin{aligned}
\frac{\partial l_{b}(\theta')}{\partial \theta} \approx\frac{\partial l_{b}(\theta')}{\partial \theta'}.
\end{aligned}
\end{equation}

After the approximation, the gradient we need to update the original parameter $\theta$ can be replaced by $\nabla_{\theta'}$, where 
\begin{equation}
\begin{aligned}
\nabla_{\theta'}=\frac{\partial l_{b}(\theta')}{\partial \theta'}.
\end{aligned}
\end{equation}
Therefore, the initial parameter $\theta$ can be directly updated using ${\nabla}_{\theta'}$ as $\theta = \theta - \beta\nabla_{\theta^{'}}$, where $\beta$ is the learning rate for the meta-test phase.

\textbf{Outer optimization} As the meta-update uses the pseudo-label which can be noisy, an outer optimization is added additionally after it. In this phase, only the data from $D_S$ with real label is used to update the model with learning rate $\gamma$, as $\theta = \theta - \gamma\nabla_\theta$.

As MAML learns the initialization of network parameters only, it can not be applied to a normal network training with a consecutive update every iteration, for the influence of the initialization can be very trivial after times of iteration. In this paper, we use the online meta-learning method \cite{onlinemeta}, and implements the procedure above in each iteration, therefore the network can benefit from the meta-learning paradigm throughout the training process.

\subsection{Text Recognition Model}
In this section, we will introduce the text recognition model we used following the flow in Section \ref{section:2}. For most of our images don't have severe deformation, the TPS module is not used in our model. During the feature extraction stage, the raw input image $X$ is fed into the CNN (we use ResNet-50 in our experiment) and generates the output feature map $F(X)=\boldsymbol{x}$, where $\boldsymbol{x}$ has the shape of $D\times1\times T$ and can be represented as $\boldsymbol{x}=\{x_1, x_2, \dots, x_t\}, x_i \subseteq R^d$. Note that $D$ and $T$ represent the channel number and the length of the feature map respectively, and the height of the feature map is set to 1. During the sequence modeling stage, we use a BiLSTM, and the hidden state of each time step can be represented as $\boldsymbol{h} = \{h_1, h_2, \dots , h_t\}, h_i \subseteq R^h$. The hidden state $\boldsymbol{h}$ is then used for the final prediction. In our model, we use an attention mechanism in \cite{attnbahdanau}. During the prediction of each time step, a context vector $c_t$ is calculated by weighting the importance of different time steps 

\begin{equation}
    c_t = \sum_{i=0}^{T}\alpha_{t,i}h_i \label{context1},
\end{equation}
where the weight $\alpha_{t,i}$ is  
\begin{equation}
    \alpha_{t,i}=\frac{exp(c_{t,i})}{\sum_{j=0}^{T}exp(c_{t,j})} \label{alpha}.
\end{equation}
The $c_{t,i}$ in Eq. \ref{alpha} is the importance of the $i$-th time step to the $t$-th time step, calculated with 
\begin{equation}
    c_{t,i}=tanh(W_{S}s_{t-1} + W_{h}h_{i}) \label{context2},
\end{equation}
where $W_s,W_h$ are the learnable parameters and $s_{t-1}$ is the hidden state of the decoder. The hidden state $s_t$ is calculated using the hidden state and the ground truth of the last time step $g_{t-1}$ (which is teacher forcing method), together with the context vector of current time step $c_t$ using a LSTM 
\begin{equation}
    s_t=LSTM(g_{t-1}, s_{t-1}, c_t) \label{decoderhidden}.
\end{equation}
Finally, a cross-entropy loss is used to calculate the classification loss on each time step
\begin{equation}
    L=\prod_{i=1}^{T}\sum_{j=1}^{k}-y_{ik}log\hat{y}_{ik} \label{textloss},
\end{equation}
where k is the size of the charset.

\section{Experiments}

\subsection{Experimental Settings}
In this section, we provide a benchmark on the dataset we proposed, and also show the experimental results and ablation studies to demonstrate the effectiveness of our method. The base text recognition model is modified on the best model in \cite{whatiswrong}. 

We implement our model using PyTorch on an NVIDIA Tesla T4. Adam is used as the outer optimizer, and SGD is used as the meta optimizer. $\alpha$ is set to $1e-3$, $\beta$ and $\gamma$ are changed during the training process. During training, we pick one domain as the target domain while the other four domains as source domains. We set the batch size to 24 per domain, which is 96 for 4 source domains and all the images are resized into $100\times32$. When using pseudo-label, we will use the training set of the target domain to generate the pseudo-label and the result is tested on the test set, which is unavailable during training.

The size of the character set in these experiments is set to 3818, which includes 3756 common Chinese characters and 62 alphanumeric characters. 

\subsection{Experimental Results}
\begin{table*}[!t]  \caption{Experiment results on five different target domains. St represents street domain; Sy represents synthetic domain; D represents documentation domain; H represents Handwritten domain; C represents car license domain}
\centering
\label{bigtabel}
\begin{tabular}{ccccccc}  
\toprule   
&St,Sy,D,H$\rightarrow$C&St,Sy,D,C$\rightarrow$H&St,Sy,C,H$\rightarrow$D&C,St,D,H$\rightarrow$Sy&C,Sy,D,H$\rightarrow$St&Average\\
\midrule   
Source Only & 22.43\% & 3.50\% &29.39\% &24.75\% &9.24\%&17.86\% \\  
MLDG \cite{metadg} &  23.85\% & 3.39\% &30.31\% &25.11\% &12.46\%&19.02\% \\     
Pseudo-Label \cite{pseudolabel} &  44.97\% & 3.77\% &51.60\% &54.11\% &15.00\%&33.89\% \\  
Meta Self-Learning (Ours) &  \textbf{58.64\%} & \textbf{5.41\%} &\textbf{64.09\%} &\textbf{65.33\%} &\textbf{16.52\%}&\textbf{42.00\%} \\  
\bottomrule  
\end{tabular}
\end{table*}

\begin{table*}[!t]  \caption{Experiment results of different settings on meta self-learning method.}
\centering
\label{smalltabel}
\begin{tabular}{ccccccc}  
\toprule   
&St,Sy,D,H$\rightarrow$C&St,Sy,D,C$\rightarrow$H&St,Sy,C,H$\rightarrow$D&C,St,D,H$\rightarrow$Sy&C,Sy,D,H$\rightarrow$St\\
\midrule   
IAOS& \textbf{58.64\%} & 4.93\% &42.94\% &37.06\% &\textbf{16.52\%}\\  
IPOA&  44.94\% & \textbf{5.41\%} &53.35\% &56.72\% &15.34\%\\     
IPOP&  41.05\% & 3.41\% &\textbf{64.09\%} &\textbf{65.33\%} &15.02\%\\  
\bottomrule  
\end{tabular}
\end{table*}

\textbf{Baseline.} The baseline model is trained with only source domains without any multi-source domain adaptation methods. The test accuracy of each domain is shown in Table \ref{bigtabel}. It can be seen that, directly using the source domain data performs badly on the target domain, which indicates that there are non-negligible domain gaps among different domains. The average accuracy among the 5 domains is 17.86\%.

\textbf{MLDG \cite{metadg}.} As described in the last section, our algorithm is a combination of meta-learning paradigm and pseudo-label method. In order to figure out the effectiveness of each part, we conduct experiments with two methods respectively. Li \etal \cite{metadg} provide a training method using the meta-learning paradigm only. During the training, the source domains are divided into meta-train set and meta-test set. The model will first update one step using the meta-train set and then validate on the meta-test set. The final model converged on source domains will be deployed on the truly held-out target domain. According to the experiment results shown in table, the MLDG is not very effective for text recognition tasks for there is only a 1.16\% improvement on average. We think the reason is that the difference between source domains and target domain is not only on appearance but also on the semantic level. For example, the document domain has a fixed length of 10 characters per sample, while other domains only contain few samples of the same length. What's more, the training data are sampled from different corpora, making it hard to learn the target domain's distribution with source domain data only.

\textbf{Pseudo-Label.} The experiment results using pseudo-label methods are shown in Table \ref{bigtabel}. As the warm-up is a necessary step for the pseudo-label method, we use the baseline model as the pre-trained model and start training using pseudo-label directly on it. For car, street, synthetic, and document domains, we set the threshold of pseudo-label confidence as 0.9. The threshold for handwritten domain is set to 0.98, for the pre-trained model performs bad on this domain. For the number of training data is very large in all domains, testing on the whole training set can be very time-consuming. Therefore, we only evaluate 50,000 images per domain, and the evaluation is done every 5,000 iterations. As shown in table, using pseudo-label can bring an up to 22.54\% gain on accuracy for a single domain, and the average accuracy increases by 16.03\%.  

\textbf{Meta Self-Learning.} Using the same setting with the pseudo-label method, the same experiments are conducted with our meta self-learning method, and the results are shown in Table \ref{bigtabel}. It can be seen that our method brings up to 13.67\% gain and 8.11\% gain on average accuracy compared with the vanilla pseudo-label method, and achieves best result on every target domain. It's worth noting that, we actually used different settings for the different target domains. It is based on the finding that for each target domain, the selection of domains for meta-update and outer optimization may affect the result greatly, and we will discuss the detail in the next section. The result shown in the table is the result corresponds to the best setting for each target domain.

\subsection{Discussion on Different Target Domains}

\begin{figure}[t]
\begin{center}
\includegraphics[height=5cm,width=7.5cm]{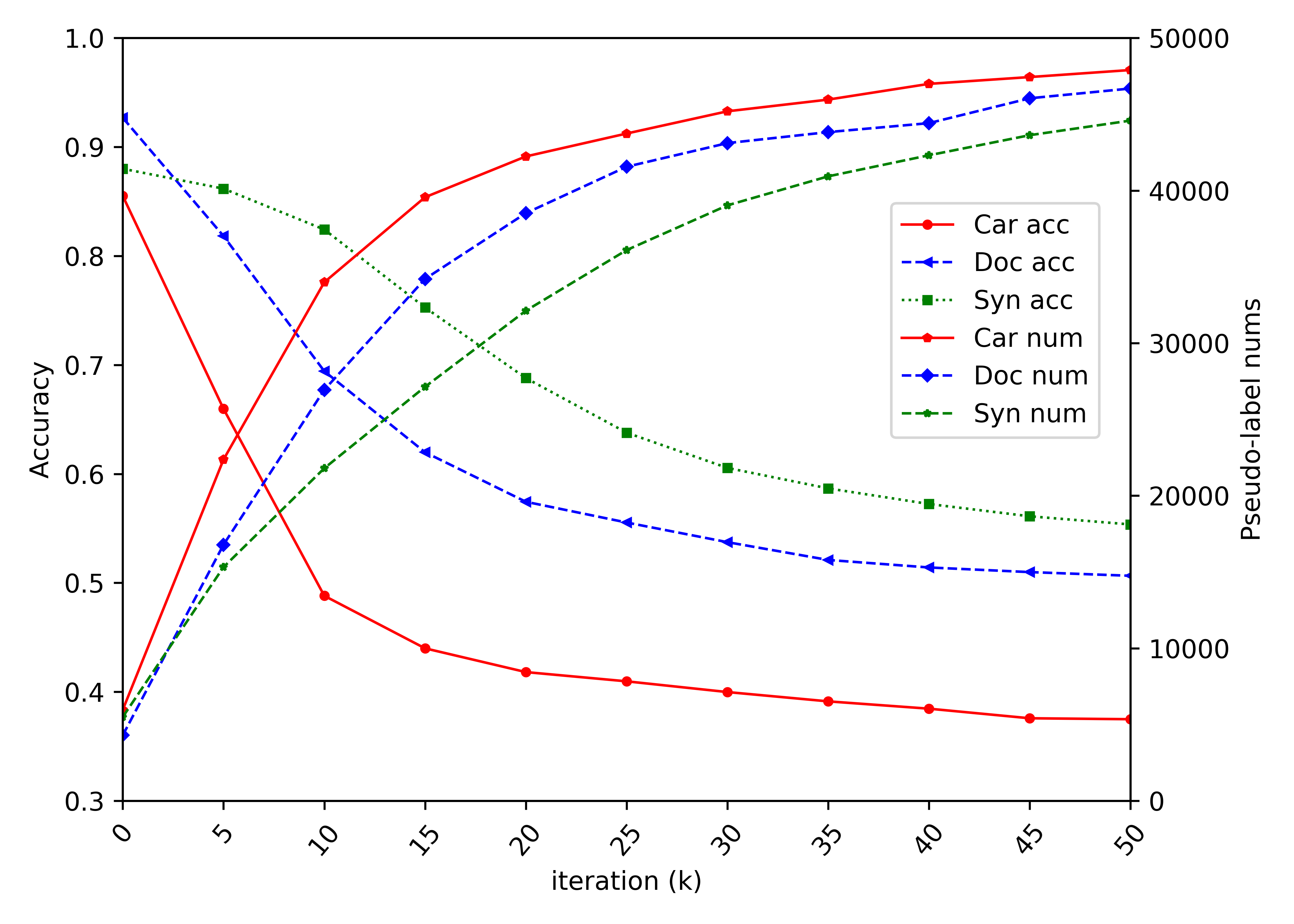}
\end{center}
\caption{The accuracy of pseudo-label for car plate, document and synthetic domain during training.}
\label{fig:3domaincompare}
\end{figure}

\begin{figure*}[t]
\begin{center}
\includegraphics[height=4cm,width=17.5cm]{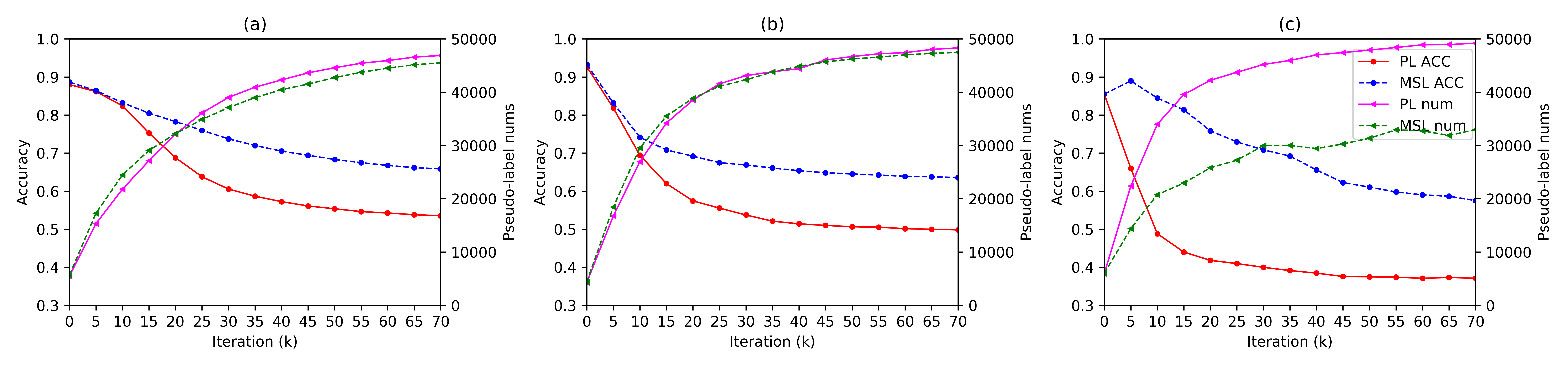}
\end{center}
\caption{The accuracy of pseudo-label during training for vanilla pseudo-label method and our method from 3 domains. (a) is the synthetic domain, (b) is the document domain, (c) is the car license domain}
\label{fig:acc}
\end{figure*}

During training, we found that the training procedure shown in Algorithm \ref{algo:1} not always performs best, therefore, we did experiments on three different settings as shown in Table \ref{smalltabel}. The main difference between these three settings is mainly reflected on the usage of pseudo-label images. 

\textbf{IAOA} represents the training procedure shown in algorithm 1, which use all 5 domains during the meta-update, and only use source domains during the outer optimization. Under this setting, we got the best results on the car license domain and the street domain. However, this setting didn't get good results on document domain and synthetic domain, and even worse than using the vanilla pseudo-label method. This may indicate that, in some domains, the use of source domain data may jeopardize the effectiveness of the pseudo-label image. Therefore, we tried to make pseudo-label play a more important role during training in some domains.

\textbf{IPOA} represents using the pseudo-label domain as meta-test set only during meta-update and use all five domains during outer optimization. In this setting, images with pseudo-label are added into the outer optimization and therefore influence the result more. It can be seen that, this setting achieves a better result than the pseudo-label method on synthetic, document and handwritten domains, which verifies our inference above. Therefore, we exploit the pseudo-label further in the next setting.

\textbf{IPOP} has the same setting with IPOA during meta-update, while only use images with pseudo-label during the outer optimization. This setting achieves great improvements on synthetic and document domains, which is 64.09\% on the document domain and 65.33\% on the synthetic domain. 

From the experiment results, we can see that the different usage of pseudo-label images can produce great gaps in the final accuracy, however, the impact is also different among different domains, but why this happens? Here we provide an intuitive explanation by the following experiments. 

During training, we record the number of generated pseudo-label images and images that have the correct pseudo-label. The results from the car plate, document and synthetic domain using vanilla pseudo-label method are shown in Fig. \ref{fig:3domaincompare}. In all three domains, the number of the generated pseudo-labels keep increasing during the training process and finally reach nearly 50,000, which is the maximum number of images we set that allowed to be used as pseudo-label. Meanwhile, the accuracy of generated pseudo-labels keep decreasing and finally converge to a value. It can be seen that the pseudo-label accuracy of the car plate domain finally converges to about 0.4 while the in document and synthetic domain, this value is between 0.5 and 0.6, which means that, the pseudo-label quality of the car plate domain is relatively low. Therefore, it is reasonable to see that the accuracy of car license domain gets lower when rely more on the pseudo-label domain, while the accuracy of document and synthetic domain get better results.

The effectiveness of our method can also be shown in Fig. \ref{fig:acc}. We demonstrate the number of pseudo-label and pseudo-label accuracy in both vanilla pseudo-label method and our methods on car, synthetic, and document domains during training. The results of synthetic domain and document domain are similar. The pseudo-label number will converge to nearly 50,000 in both two methods, while our method stably gets a higher accuracy on the generated pseudo-label, which is on average 10\% higher than the vanilla pseudo-label method. For the car plate domain, the number of pseudo-label in our method is controlled to about 30,000, and get a 20\% promotion on the accuracy. These indicate that our method can produce pseudo-label with higher quality and get better training results. 

\subsection{Application Analysis}
Our method provides a self-learning method for multi-source domain adaptation problems, but it can also be transferred into the single-source domain adaptation problem, where $D_S$ contains only one domain. Actually, our method is a self-learning framework and is model-agnostic, therefore can be easily applied to any task using self-learning methods. However, as discussed in the last section, the paradigm for different tasks may need to be changed according to the quality of pseudo-labels. In future work, we will try to find a theoretical explanation and a unified framework for our method.

\section{Conclusion}
In this paper, we collect and generate a multi-source domain adaptation dataset for text recognition. To our best knowledge, this is the first and the largest publicly available dataset for this area, which is very meaningful for the great significance of both domain adaptation and text recognition problems. We also propose a new meta self-learning method for the multi-source domain adaptation problem, which is model-agnostic and can be easily applied to other tasks. Extensive experiments are done on our dataset to provide a benchmark and demonstrate the effectiveness of our method. However, our dataset is still very challenging because of the large scale of charset and notable domain shift among domains, and worth more exploration in the future.

\section*{Acknowledgements}
This work was supported in part by 111 Project of China (B17007), and in part by the National Natural Science Foundation of China (61602011).
\newpage

{\small
\bibliographystyle{ieee_fullname}
\bibliography{egpaper_arXiv}
}

\end{document}